\pdfoutput=1
\documentclass[letterpaper]{article} % DO NOT CHANGE THIS
\usepackage{aaai24}  % DO NOT CHANGE THIS
\usepackage{times}  % DO NOT CHANGE THIS
\usepackage{helvet}  % DO NOT CHANGE THIS
\usepackage{courier}  % DO NOT CHANGE THIS
\usepackage[hyphens]{url}  % DO NOT CHANGE THIS
\usepackage{graphicx} % DO NOT CHANGE THIS
\graphicspath{ {./img/} }
\usepackage{natbib}  % DO NOT CHANGE THIS AND DO NOT ADD ANY OPTIONS TO IT
\usepackage{caption} % DO NOT CHANGE THIS AND DO NOT ADD ANY OPTIONS TO IT
\usepackage{algorithm}
\usepackage{algorithmic}

\usepackage{newfloat}
\usepackage{listings}
\DeclareCaptionStyle{ruled}{labelfont=normalfont,labelsep=colon,strut=off} % DO NOT CHANGE THIS
\floatstyle{ruled}
\newfloat{listing}{tb}{lst}{}
\floatname{listing}{Listing}
\usepackage{enumitem}
\usepackage{nameref}

\title{Artificial Intelligence Approaches for Energy Efficiency: A Review}
\author{
    Alberto Pasqualetto\textsuperscript{\rm 1}\equalcontrib,
    Lorenzo Serafini\textsuperscript{\rm 2}\equalcontrib,
    Michele Sprocatti\textsuperscript{\rm 3}\equalcontrib\\
}
\affiliations{
    Department of Information Engineering, University of Padua\\
    Via Gradenigo 6, 35131 Padova, Italy\\
    \{alberto.pasqualetto.2\textsuperscript{\rm 1}, lorenzo.serafini.1\textsuperscript{\rm 2}, michele.sprocatti\textsuperscript{\rm 3}\}@studenti.unipd.it
}

\nocopyright

\begin{document}

\maketitle

\begin{abstract}
\citeauthor{unitednationsSustainableDevelopmentGoals2015} set Sustainable Development Goals and this paper focuses on 7th (Affordable and Clean Energy), 9th (Industries, Innovation and Infrastructure), and 13th (Climate Action) goals.
Climate change is a major concern in our society; for this reason, a current global objective is to reduce energy waste.
This work summarizes all main approaches towards energy efficiency using Artificial Intelligence with a particular focus on multi-agent systems to create smart buildings.
It mentions the tight relationship between AI, especially IoT, and Big Data.
It also explains the application of AI to anomaly detection in smart buildings and a possible classification of Intelligent Energy Management Systems: Direct and Indirect.
Finally, some drawbacks of AI approaches and some possible future research focuses are proposed.
\end{abstract}

\section{Introduction} \label{sec:introduction}
According to estimates by the United States Energy Information Administration, 40\% of the annual CO\textsubscript{2} emissions are directly related to electricity consumption.

\begin{figure}[h]
    \centering
    \includegraphics[width=0.45\textwidth]{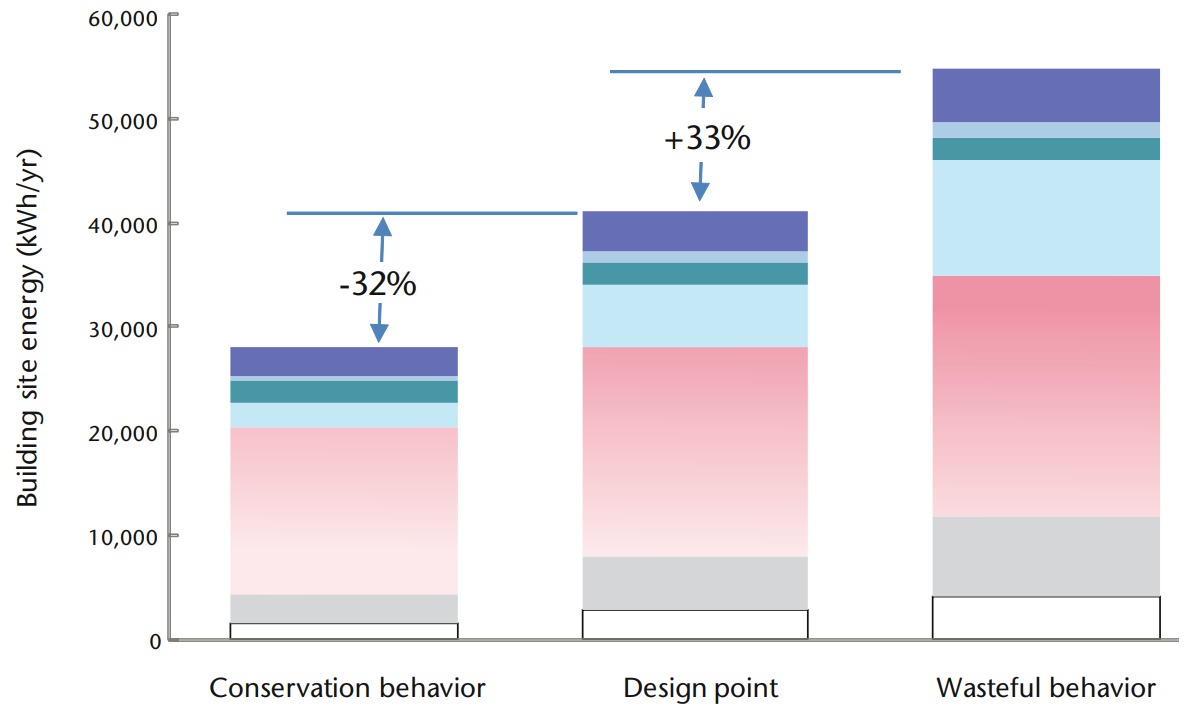}
    \caption{Conservation vs wasteful behavior in energy consumption of a building according to World Business Council For Sustainable Development, colors represent different energy usage types, image from \cite{WBCSDenergyefficiencybuildings2009}.}
    \label{fig:conservation_vs_wasteful_consumption}
\end{figure}

A way to tackle pollutant emissions is to reduce the amount of wasted energy.
Thanks to a more efficient use of energy, we can reduce energy consumption by more than 30\% as shown in Figure \ref{fig:conservation_vs_wasteful_consumption}.

The most energy-consuming part of a building is the Heating, Ventilation, and Air Conditioning (HVAC) system \cite{WBCSDenergyefficiencybuildings2009}, for this reason, a special focus is put on this system in the literature.

To improve awareness about energy waste habits and reduce it, consumers must first monitor their energy consumption and manage it after receiving appropriate advice \cite{darbyEffectivenessfeedback2006}.
This can be achieved practically by using an Energy Management System (EMS). Many approaches use Artificial Intelligence (AI) tools to design Intelligent Energy Management Systems (IEMS), detect and minimize faults and increase energy efficiency. Here we review some possible approaches found in the literature.

AI in IEMS has a strong coupling with data collection from sensors, in this case, due to the amount of elaborated data, we can talk about "Big Data".

\section{Big Data Usage in IEMS} \label{sec:big_data}
Big Data (BD) refers to huge amounts of data, both structured and unstructured, that can be used to extract useful information.
BD usage has been increasing in recent years, and it is expected to continue to grow in the future because of the massive increase in global data creation\cite{IDCStatistaVolumeData2021}.

BD is characterized by the V's (Volume, Velocity, Variety, Veracity, and Value, and even more are being added to the list during the years) \cite{sagirogluBigDataReview2013} and, regarding energy efficiency, the most relevant characteristics are \cite{koselevaBigDataBuilding2017,zhouBigDataDriven2016}:
\begin{description}
    \item [Volume] The amount of electricity consumption data collected by 1 million smart meters in 1 year (collecting once every 15 minutes) is about 3 TB; this presents a challenge for data storage and analysis.
    \item [Velocity] To analyze the data in real-time, the data processing speed must be fast enough to keep up with the incoming information, which means in a range that goes from sub-second to 15 minutes.
    \item [Variety] The data collected from smart meters is heterogeneous, including electricity consumption data, weather data, and customer information: each of them is useful in a different part of the industry. Also, this data is structured, semi-structured, and unstructured.
\end{description}

BD's primary focus is on collecting data and gaining insights, while AI aims to accomplish tasks.

The usage of BD for energy-efficient buildings is powered by the Internet of Things (IoT), the collection of local data from sensors and other services that provide local and global weather data, building information, and historical data provided by societies, groups, and commercial companies. It is also worth mentioning the potential usage of 3D topographic data which can help to assess the shadowing of surroundings, air flow, noise level, and other site-specific information even before the construction of a building  \cite{mehmoodReviewAIBD2019}.

% One perk of using BD can be reducing the data collection frequency to save energy, disk space, and processing time, while still providing good results based on the site-specific data collected previously \cite{mehmoodReviewAIBD2019}.

Once the data is collected, it can be used to train (fine-tune) AI models to predict weather and inhabitant patterns together with the building's characteristics like insulation, heat transfer, household appliances energy consumption, etc. \cite{mehmoodReviewAIBD2019}.

Big Data can not only be used commercially, or for personal use, but also for research purposes thanks to the massive amount of geographically distributed data, which can be used for weather study and prediction or for AI training, etc..

\section{Artificial Intelligence for Energy Efficiency} \label{sec:AI}
The most promising use of AI is in smart buildings equipped with sensors and actuators.
They can be components of a smart grid that uses a Demand Side Management system (DSM) to plan, control, and manage the loads of electricity.

% A possible application of AI to an energy management system is the Demand-Response programs (DRPs). A DRP system can be classified into time-based rate (TBR) or incentive-based programs (IBPs). The main difference between the two systems is that with TBR we propose an optimal scheduling of energy consumption to minimize the payment and the appliance waiting time based on the user's lifestyle, while IBP requires the user to contribute voluntarily to a pattern where the system operator turn off some appliances during peak time.

AI systems can also be used for a statistical approach in order to make predictions about consumer energy use; the main 4 steps are:
\begin{enumerate}
    \item Data collection
    \item Data preprocessing
    \item Model training
    \item Model testing
\end{enumerate}

\citeauthor{farzanehArtificialIntelligenceEvolution2021} presented several AI techniques and models with their respective use cases:
\begin{description}
    \item [Decision trees] to predict blackouts and to define energy storage planning and management in buildings.
    \item [Random forest] to predict energy consumption.
    \item [Wavelet neural networks] to predict time-series data for hybrid renewable residential micro-grids.
    \item [Naive Bayes models] to solve energy problems in buildings when having a prior probability, and the next consume guess is needed.
    \item [Artificial neural networks] to evaluate thermal comfort and to give suggestions to the user to save energy.
    \item [Deep learning] to estimate the power consumption of appliances in a distributed system.
    \item [Regression] to predict monthly heating and cooling demands.
    \item [Genetic algorithms] to minimize the total energy cost in a dynamic pricing scheme and it can be also used for finding the optimal energy consumption to provide a comfortable temperature.
    \item [Fuzzy logic] to forecast the demand load based on changes in indoor and outdoor conditions.
    \item [Particle swarm optimization] together with other techniques to find the optimal solution in problems like estimating the energy demand.
    \item [K-nearest neighbor] to analyze energy demands.
    \item [Principal Component Analysis (PCA)] to reduce the dimensionality of the data and then use another AI method to make the prediction.
\end{description}
Different models can be combined to make better predictions.

% After this the paper talks about some opportunities in which an AI system can be applied, these include:
% \begin{enumerate}
%     \item Energy efficiency: reduce energy consumption during peak hours, detect equipment failures before they occur, to also enable district
%     heating/cooling, and to help efficient design of the buildings.
%     \item Renewable energy forecasting: predict energy production given the meteorological data.
%     \item Energy accessibility: manage electricity grids and improving the accessibility of renewable energy sources.
% \end{enumerate}

\subsection{Multi-Agent Systems} \label{ssec:MAS}
An agent is a computer system situated in some environment, and that is capable of autonomous action in such environment to meet its design objectives \cite{wooldridgeIntelligentAgents1999}. It can be described as the implementation of the "Belief-Desire-Intention" (BDI) model, where the agent has beliefs about the environment, desires to achieve some goals, and intentions to act to achieve them.
% autonomous, which means without the intervention of any other agent,

A Multi-Agent System (MAS) consists of an environment, objects, agents, and all their relations, a set of actions that can be performed by the entities to change the environment \cite{ferberMultiAgentSystemsIntroduction1999}; each agent cooperates with the others to achieve a common goal (but it can also have its own objectives).
This system makes it possible to emulate the actions and interactions of human organizations through the decomposition of problems into smaller parts; problems are solved by agents communicating.
An example of decomposition occurs when a group of agents monitor the comfort temperature set by the user and another group controls the HVAC system and communicates with a service agent that collects past data to help select a new ideal temperature for the residents.

% The granting of learning capabilities to agents has allowed MASs to reason about different possibilities (different configurations of lighting, heating, cooling or ventilation in an Intelligent Building (IB), for example), autonomous decision-making, learning new situations and errors from the past (a temperature configuration has been made that allows the maximum energetic optimization according to the set of rules, but the user has decided to increase the temperature a little), programming of tasks and forecasting of future situations.

% MAS is usually developed by exploiting Object-Oriented Programming to express the BDI architecture; on top of that are built the communication, cooperation, and coordination capabilities.
MASs often rely on frameworks that provide the necessary tools to create and manage agents, such as JADE (Java Agent DEvelopment framework) \cite{bellifemineJADE1999} and PADE (Python Agent DEvelopment framework), the most used in the MAS community. Both of them are compatible with the FIPA (Foundation for Intelligent Physical Agents) \cite{obrienFIPA1998} standard, which defines the communication protocols between agents.

In the past there has been a lack of standardization in the communication protocols; nowadays BACnet is the leader in the market for HVAC \cite{BACnetWebsite} systems; in the last years an open-source protocol supported by big players in the market, Matter, is arising and being established as the new standard for all the IoT devices \cite{matterWebsite,matterGithub}.
Optimization algorithms are used to lower the energy consumption of the building, two types of algorithms can be used: exact or heuristic. The first ones are based on mathematical models and can find the optimal solution; the second ones are not necessarily optimal, but they are faster and overall they are better suitable in problems with a human in the loop, like the energy efficiency of a building, in which a trade-off with users' comfort is needed. MASs have been proven to be promising as heuristic techniques for solving such problems whose domains are distributed, complex, and heterogeneous.

An interesting agent learning approach is the one described by \citeauthor{tuylsEvolutionaryGameTheory2005,tuylsEvolutionaryDynamicalAnalysis2006} who use Reinforcement Learning (RL): the inhabitants of an Intelligent Building (IB) are modeled as agents that play a game in which they obtain a benefit if they lower their energy consumption and they are punished if they increase it. Agents acting as humans learn to optimize the trade-off between energy consumption and their comfort. In this way, the system can learn the best configuration for the IB.

MAS usage in the energy efficiency field can be divided into three main areas of application \cite{gonzalezMultiAgentSystemsApplications2018}:
\begin{description}
    \item [Demand Response (DR)] the ability of the supplier-client system to react to the supplier's request to adjust energy consumption according to available supplies or to reduce the cost of energy. This is particularly useful in the case of demand peaks, failure, or maintenance in the grid or to exploit renewable energy sources at their best.
    % A review of several DR's works can be found in tables 1,2 in \citeauthor{gonzalezMultiAgentSystemsApplications2018}.
    \item [Human Behavior Simulation] the simulation ability, which is, as already mentioned, a strong potential of MASs. Simulations focus on humans present in the house: their habits and preferences, their interactions with the building (appliances, heating \& cooling, lighting, etc.), and the consequences of their actions on energy consumption also taking into account all the external factors (weather, time of the day, price of energy, etc.).
    Each person can be modeled by a separate agent, allowing the evaluation of a different behavior for each inhabitant, taking into account the heterogeneity of human behavior in the comfort pursuit.
    % in fact \citeauthor{hagrasHierarchicalFuzzyGenetic2003} represented each person with a fuzzy-genetic agent in their work.
    The system aims to optimize energy consumption without reducing the user's comfort sensation.
    \citeauthor{yangDevelopmentMultiagentSystem2013} studied how the user's preferences can be perceived (asking the users or learning from manual adjustments made by users).
    Main parameters that influence energy consumption and need, a fundamental stage for improving optimization models, can be found using Principal Component Analysis (PCA) \cite{darakdjianDataMiningBuilding2019}.
    A review of other works can be found in tables 3,4 in \citeauthor{gonzalezMultiAgentSystemsApplications2018}.
    \item [Wireless Sensor Networks (WSN) management] the use of MASs to simulate real buildings and how sensors would act in it.
    A classification of the most relevant elements to be monitored: occupation, preferences, activities, temperature, etc. can be found in \citeauthor{nguyenEnergyIntelligentBuildings2013}.
    The integration of Real-Time Localization Systems boosted MAS systems evolution in this field, achieving an average energy saving of 17-22\% \cite{gonzalezMultiAgentSystemsApplications2018}.
    % A review of works about WSN management can be found in tables 5,6 in \citeauthor{gonzalezMultiAgentSystemsApplications2018}.
    After a WSN simulation, which is especially useful in the research field, a real WSN can be deployed in situ.
\end{description}

Summing up a MAS-based system should focus on the following aspects \cite{gonzalezMultiAgentSystemsApplications2018}:
\begin{description}[style=sameline]
    \item [Knowledge and learning] the agents should know the user's preferences which affect energy consumption such as tastes in terms of comfort, habits, timetables, etc., also agents should know the environment characteristics like the building's structure, the thermal relation with the external environment, weather forecast, etc.
    \item [Analysis, adaptation, and communication with the environment] the agents should be able to analyze their prior knowledge and communicate the information that the MAS must use to adapt to the environment.
    To communicate with the environment, and send and receive information from it, it is necessary to deploy a WSN.
    \item [Decision-making] the agents should be able to make decisions based on the information they have, the environment, and the user's preferences. Here algorithms should give greater weight to the factors they consider appropriate; to identify the most important ones, a PCA can be used \cite{darakdjianDataMiningBuilding2019}.
\end{description}

The results in energy saving that can be achieved with MAS varies between 17\% and 41\% based on the specialization and the target human satisfaction of the system \cite{gonzalezAgreementTechnologiesEnergy2018,spearsFormalAnalysisPotential2005}

\section{Anomaly Detection} \label{sec:anomaly_detection}
AI can be used to detect anomalous usage of electricity and its causes \cite{himeurArtificialIntelligenceBased2021}.
Anomalies can be caused by different factors like:
\begin{itemize}
    \item Wrong behavior of the user that forgets to turn off some electrical appliances.
    \item Some appliances not working like they are supposed to (faults).
    \item Electricity thefts.
\end{itemize}
Moreover, an anomaly must be detected depending on the time of the day or the season, for example, it is not an anomaly if the heating system is on during the winter, but it is during the summer.

To determine if there is an anomalous electricity usage, the AI model must take care of a lot of aspects like the power consumption but also appliance-specific parameters, occupancy patterns, and weather conditions.

Many anomaly detection approaches can be found in the literature:
\begin{itemize}
    \item \textit{Unsupervised detection} (clustering, dimensionality reduction)
    \item \textit{Supervised detection} (Neural Networks, regression, probabilistic models)
    \item \textit{Ensemble method} (boosting, bagging)
    \item \textit{Feature extraction} (distance-based, time-series analysis, etc.) % density-based, graph-based
    \item \textit{Hybrid learning}
    \item \textit{Other techniques}
\end{itemize}

Also, systems can be categorized based on the anomaly detection level:
\begin{description}
    \item [Aggregated level] detect anomalous consumption based on data of the main building-wise meter but, doing so, the information about the responsible appliance is lost.
    \item [Appliance level] detect using individual sub-meters to have fine-grained tracking of abnormalities.
    \item [Spatial-temporal level] detect abnormal usage related to specific days and hours and, possibly, provide users some feedback to reduce energy waste.
\end{description}

A relevant challenge in this field is defining what exactly is an anomaly: it can be described as the event that occurs when the power usage is an outlier, both regarding a peak in power usage or a too-long electricity consumption; but this definition doesn't take in account the fact that if an appliance is running for a long time-span, for example, long cooking in the oven, then it can also not be an anomaly.
The aim is the production of systems that have good accuracy and consider all the possible abnormal usage.

A current problem is the lack of annotated datasets: the existent ones are mostly not labeled or contain only a small amount of anomaly samples, because of the rarity of anomalies in reality.

\section{Direct vs Indirect Control} \label{sec:direct_indirect_control}
\citeauthor{mischosIntelligentEnergyManagement2023} introduced a new distinction in IEMS classification:
\begin{description}
    \item [Direct Control IEMS (DCIEMS)] includes all systems with the ability to modify the environment using actuators; these systems are able to process data, make decisions, and execute them, without the intervention of a human being.
    The advantages of Direct Control IEMS are that the occupants don't need to alter their daily routine and these systems can be involved in fault detection. Furthermore, they are better suited for people with reduced mobility.
    %  like a physical disability or elderly people.
    \item [Indirect Control IEMS (ICIEMS)] instead puts humans in the position of the actuator, forming a human-in-the-loop architecture.
    These systems aim to change the behavior of end-users to stop energy-wasting habits. To achieve this goal, suggestions are provided to the users through interfaces.
    This approach is cheaper since there is no need to buy actuators like smart sockets for every appliance; moreover, by inducing an eco-friendly behavior in the users, they will apply the acquired habits in all the other aspects of their lives, like at their workplace.
\end{description}

Both the IEMS types produce a lot of data: Direct Control IEMS through sensors, and Indirect Control IEMS through feedback from end users.
AI tools can be used to process this quantity of data as described in section "\nameref{sec:big_data}".
Machine learning tools can extract from such data the behavioral usage patterns of the users, predicting their future needs and personalizing the recommendations for each one of them.
Deep learning tools can monitor the energy profile of the environment and derive information about the energy consumption of appliances.
They are also able to make predictions about the next consumption in a determined time window.

\section{Future proposals} \label{sec:future}
Until today only a few studies have been conducted on the use of Deep Learning in the energy efficiency field, in the authors' opinion this should be the focus of future research.
A possible approach could be to combine Deep Learning with Reinforcement Learning to enhance the learning done by Deep Learning Neural Networks in DCIEMS through both the refinements provided by a reward function based on the environment changes and the feedback provided by the human.

Although the aforementioned proposal, the authors recognize that a major issue in the implementation of DL systems is the computational power requirements that are needed in training/fine-tuning, but also at inference, such power is not always available in the buildings due to the high cost of the required hardware.

Using DL, the "Explainability" issue arises, since the user must understand the actions (DCIEMS) or suggestions (ICIEMS) from the system in order to be compliant with them. An optimal system should be able to explain the reasons behind its output in a way that the user can understand.

Another proposal could be to implement gamification experiences in ICIEMS architecture to engage users in the energy-saving process, making them more aware of their energy consumption and more willing to change their habits both at home and in other buildings.
An example implementation could be a mobile application that rewards the user with points for every energy-saving action taken; the user can then compete with other users globally or locally against occupants of the same building.

\section{Conclusion} \label{sec:conclusion}
This paper has reviewed the literature on the use of AI and its different techniques in the energy efficiency field, focusing on Big Data produced by IoT, Anomaly Detection with AI, and the different approaches in interaction with humans and buildings.

Energy efficiency is a very important topic in current society since, as many studies pointed out, it can lead to a reduction in CO\textsubscript{2} emissions and, consequently, to a reduction of climate change.
This also reflects on people's health: \citeauthor{macnaughtonEnergySavingsEmission2018} found that for every dollar saved on energy expenditures by smart buildings, 77 cents are saved in health and climate benefits.

All the reviewed papers pointed out problems in AI systems, especially the privacy aspects because models know a lot of information about the users and the buildings, and such information can be used to harm inhabitants; in case of data leaks malicious people can harm users by knowing if they are at home.
A way of enforcing users' privacy in global models training is using anonymized data or, even better, using federated learning which helps to train algorithms over various de-centralized edge servers without sharing data.

In general, these systems are not even immune to attacks: in particular, DCIEMS are sensitive to attacks since they rely on IoT devices; a possible attack can be Denial of Service (DoS) and False Data Injection (FDI), which can affect electricity bills and load consumption.

\bibliography{references}

\end{document}